\documentclass{article}

\usepackage{microtype}
\usepackage{graphicx}
\usepackage{subfigure}
\usepackage{booktabs} 
\usepackage{diagbox}
\usepackage{bbm}
\usepackage{multirow}

\usepackage{hyperref}

\usepackage[accepted]{icml2025}

\usepackage{amsmath}
\usepackage{amssymb}
\usepackage{mathtools}
\usepackage{amsthm}

\usepackage[capitalize,noabbrev]{cleveref}

\theoremstyle{plain}

\theoremstyle{definition}

\theoremstyle{remark}

\usepackage[textsize=tiny]{todonotes}

\icmltitlerunning{Offline Reinforcement Learning with Discrete Diffusion Skills}

\begin{document}

\twocolumn[
\icmltitle{Offline Reinforcement Learning with Discrete Diffusion Skills}



\icmlsetsymbol{equal}{*}

\begin{icmlauthorlist}
\icmlauthor{Ruixi Qiao}{casia,ucas}
\icmlauthor{Jie Cheng}{casia,ucas}
\icmlauthor{Xingyuan Dai}{casia}
\icmlauthor{Yonglin Tian}{casia}
\icmlauthor{Yisheng Lv}{casia,ucas}
\end{icmlauthorlist}

\icmlaffiliation{casia}{State Key Laboratory of Multimodal Artificial Intelligence Systems, Institute of Automation, Chinese Academy of Sciences}
\icmlaffiliation{ucas}{School of Artificial Intelligence, University of Chinese Academy of Sciences}

\icmlcorrespondingauthor{Yisheng Lv}{yisheng.lv@ia.ac.cn}

\icmlkeywords{Machine Learning, ICML}

\vskip 0.3in
]




\begin{abstract}
Skills have been introduced to offline reinforcement learning (RL) as temporal abstractions to tackle complex, long-horizon tasks, promoting consistent behavior and enabling meaningful exploration. While skills in offline RL are predominantly modeled within a continuous latent space, the potential of discrete skill spaces remains largely underexplored. In this paper, we propose a compact discrete skill space for offline RL tasks supported by state-of-the-art transformer-based encoder and diffusion-based decoder. Coupled with a high-level policy trained via offline RL techniques, our method establishes a hierarchical RL framework where the trained diffusion decoder plays a pivotal role. Empirical evaluations show that the proposed algorithm, Discrete Diffusion Skill (DDS), is a powerful offline RL method. DDS performs competitively on Locomotion and Kitchen tasks and excels on long-horizon tasks, achieving at least a 12 percent improvement on AntMaze-v2 benchmarks compared to existing offline RL approaches. Furthermore, DDS offers improved interpretability, training stability, and online exploration compared to previous skill-based methods.
\end{abstract}

\section{Introduction}
Offline reinforcement learning (RL) \cite{levine2020offline} aims at learning effective policies from historical interaction data without the need to interact with the real environment, making it particularly valuable for expensive or high-risk applications such as robotic manipulation and autonomous driving. However, applying offline RL algorithms off-the-shelf \cite{fujimoto2021minimalist, kumar2020conservative, kostrikov2021offline} to long-horizon, sparse-reward tasks remains a significant challenge. Some drew inspiration from the fact that humans typically solve complex problems by first acquiring a repertoire of skills and then logically combining these skills to address more intricate challenges. Therefore, the concept of skill has been introduced into offline RL as temporal abstractions for their potential to enhance long-horizon planning and exploration capabilities \cite{chen2024scar, li2024skills, mete2024quest, eysenbach2018diversity}. 

Skill-based methods \cite{sharma2019dynamics, ajay2020opal} have used VAE-like architectures to extract skills encoded as continuous latent embeddings from offline datasets. The trained decoder is then used as the low-level policy of a hierarchical RL framework, where the high-level policy selects a skill, and the low-level policy executes it \cite{ajay2020opal, venkatraman2023reasoning}. Although such continuous skill spaces provide a diverse range of skills, the interpretability of individual skills remains elusive. Furthermore, previous works used simple network structures for the encoder and the decoder that may not be able to fully capture the complex state-action distribution.

To enhance the interpretability of skill-based methods and facilitate downstream learning, we propose modeling skills within a discrete skill space, enabling each skill to be associated with a distinct semantic meaning. We leverage state-of-the-art network architectures such as Transformers \cite{vaswani2017attention} and diffusion models \cite{ho2020denoising} to efficiently learn skills in a discrete space. Because these architectures have demonstrated great capability to model complex distributions with their success in large language models \cite{achiam2023gpt, bai2023qwen} and image/video generation \cite{wang2022diffusion}. Specifically, we employ a transformer-based encoder to map state-action sequences into compact skill representations and a diffusion-based decoder to generate actions conditioned on skill and state. Afterwards, the offline dataset is relabeled with acquired skills for high-level policy learning with Implicit Q-Learning (IQL) \cite{kostrikov2021offline}. The high-level policy and diffusion decoder forms DDS agent with a hierarchical RL architecture.

Through experiments, we demonstrate that discrete skill corresponds to distinct behaviors and induces meaningful interactions with the state. Ablation experiments highlight the critical role of network architecture in achieving effective skill learning. Moreover, by using a discrete skill space, the action space is discretized to facilitate high-level policy learning \cite{luo2023action, vieillard2021implicitly} and online exploration.

In summary, our work has three main contributions. Firstly, we present DDS, a capable and robust offline RL method that models skills in a compact discrete skill space utilizing state-of-the-art network architectures. Secondly, discrete skills enhance the interpretability of skill-based methods by providing a clearer understanding of the agent's behavior. Lastly, discrete skills can be effectively leveraged to improve online RL performance by accelerating policy learning and enabling more efficient exploration in complex sparse reward environments.

\section{Related Work}

\textbf{Offline RL} Offline RL focuses on training an agent solely from collected datasets. A key challenge in this setting is the extrapolation error, which means that the values of actions not represented in the dataset are inaccurately estimated. This could significantly degrade the performance of the RL agent. Existing approaches to address this issue can be broadly categorized into two groups. The first group introduces conservatism into value estimation or policy learning, as seen in methods like CQL \cite{kumar2020conservative}, BEAR \cite{kumar2019stabilizing}, and IQL \cite{kostrikov2021offline}. The second group frames offline policy learning as a conditional generative process, exemplified by works such as Decision Transformer (DT) \cite{chen2021decision} and Decision Diffuser (DD) \cite{ajay2022conditional}. Our work bridges these two paradigms by leveraging a diffusion-based generative model as the low-level policy and employing offline Q-learning for the high-level policy.

\textbf{Self-supervised skill discovery} Skill has been introduced to offline RL as temporal abstraction of actions or state-action sequences \cite{pertsch2021accelerating, tasseskill, hao2024skill}. These methods typically employ an encoder-decoder architecture with prior regularization, akin to the $\beta$-VAE framework \cite{higgins2017beta}. The encoder maps the recorded sequences to a continuous latent skill space, regularized by a prior network, while the decoder reconstructs the action sequence conditioned on the skill and states \cite{venkatraman2023reasoning}. In a hierarchical RL framework, the skill decoder generally functions as the low-level policy, guided by a high-level policy that learns to direct the low-level policy to solve tasks via reinforcement learning.

Our work distinguishes itself from prior approaches in two key aspects: first, we employ a compact discrete skill space, and second, our framework does not require regularization with a prior network due to the intrinsic regularization properties of the discrete space. Although discrete skills have been explored in prior works \cite{peng2019mcp, bacon2017option}, the skills proposed in our work are specifically designed for the offline setting and feature a distinct interaction with the state space to generate actions.

\textbf{Diffusion Policies} have emerged as a powerful class of expressive policies for RL, inspired by the remarkable success of diffusion models in image and video generation \cite{saharia2022photorealistic, ho2022imagen}. \citet{wang2022diffusion} was the first to integrate diffusion policies into the Q-learning framework. Subsequently, \citet{chen2022offline} proposed decoupling policy learning into behavior learning and action evaluation, allowing greater flexibility in the choice of guidance mechanisms.

In contrast to these approaches, we employ diffusion models as low-level policies, leveraging mainly their conditioned generation capabilities and avoiding value training with diffusion models to significantly accelerate the training process. While \citet{kim2024robust} and \citet{liang2024skilldiffuser} also utilize diffusion decoders, their focus lies primarily on the multi-task setting rather than the offline RL domain.

\section{Preliminaries}
\subsection{Diffusion Probabilistic Model}
The Diffusion Probabilistic Model (DPM) is a generative model that learns to generate data by modeling a series of transformations between a complex data distribution (e.g. images, videos) and a simple prior distribution (e.g. Gaussian noise) \cite{sohl2015deep}. The process consists of two key steps: the forward noising process and the reverse denoising process.

In the forward noising process, Gaussian noise is added to the data through a fixed Markov chain according to a variance schedule $\beta_{1},...,\beta_{T}$: 
\begin{align}
q( x_{1:T} | x_{0} )= \prod_{t=1}^{T} q ( x_{t} | x_{t-1} ), \label{eq:fnp1} \\
q ( x_{t} | x_{t-1} )= \mathcal{N}(x_{t};x_{t-1}\sqrt{1-\beta_{t}},\beta_{t}\mathrm{I} ),\label{eq:fnp2}
\end{align}
where $x_{0},...,x_{T}$ are latent variables and $\mathcal{N}$ denotes a Gaussian distribution. Let $\alpha_{t} = 1 - \beta_{t}$ and $\bar{\alpha_{t}} =  {\textstyle \prod_{i=1}^{t}}\alpha_{i}$ The forward distribution can be computed for a timestep $t$ in closed form: $q ( x_{t} | x_{0} ) = \mathcal{N}(x_{t};\sqrt{\bar{\alpha_{t} }} x_{0},(1-\bar{\alpha_{t}})\mathrm{I})$.

In the reverse denoising process, the model learns to reverse the forward process. Starting from pure noise, it iteratively removes noise to reconstruct the data:
\begin{align}
p_{\psi } (x_{0:T})=p(x_{T})\prod_{t=1}^{T} p_{\psi } (x_{t-1}|x_{t}), \label{eq:rdp1} \\
p_{\psi } (x_{t-1}|x_{t}) = \mathcal{N}(x_{t-1};\mu_{\psi }(x_{t} ,t),\Sigma_{\psi }(x_{t} ,t) ),\label{eq:rdp2}
\end{align}
where the target distribution is $p_{\psi } (x)$. The reverse process is trained by minimizing a surrogate loss-function \cite{ho2020denoising}:
\begin{equation}
\mathcal{L}(\psi) = \mathbb{E}_{t\sim [1,T],x_{0}\sim q(x_{0}),\epsilon \sim \mathcal{N(\mathrm{0},\mathrm{I})}}||\epsilon -\epsilon _{\psi }(x_{t},t )||_{2}, \label{eq:diffloss}
\end{equation}
$\epsilon _{\psi }$ is usually parameterized through deep neural networks.

\subsection{Implicit Q-Learning}
Implicit Q-learning is an offline RL algorithm that approximates the expectile $\tau_{IQL}$ over the distribution of actions \cite{kostrikov2021offline}. For critic $Q_{\theta}(s,a)$, target critic $Q_{\hat{\theta}}(s,a)$, and value network $V_{\varphi }(s) $, the value object is:
\begin{align}
\mathcal{L}_{V}(\varphi) = \mathbb{E}_{(s,a)\sim D}[L_{2}^{\tau_{IQL} }(Q_{\hat{\theta}}(s,a)-V_{\varphi }(s)) ], \label{eq:iql1} \\
L_{2}^{\tau_{IQL} }(u)= |\tau_{IQL}-\mathbbm{1}(u<0)|u^{2}. \label{eq:iql2}
\end{align}
This value function is used to update the Q-function:
\begin{equation}
\mathcal{L}_{Q}(\theta) = \mathbb{E}_{(s,a,s^{'} )\sim D}[r(s,a)+\gamma V_{\varphi }(s^{'})-Q_{\theta}(s,a)]. \label{eq:iqlloss}
\end{equation}
For policy extraction, IQL uses AWR \cite{peters2007reinforcement}, which trains the policy via weighted regression by minimizing:
\begin{equation}
\mathcal{L}_{\pi }(\phi ) = \mathbb{E}_{(s,a)\sim D}[\mathrm{exp}(\alpha (Q_{\hat{\theta}}(s,a)-V_{\varphi }(s)))\mathrm{log}\pi_{\phi }(a|s) ], \label{eq:awrloss}
\end{equation}
where $\alpha$ is a non-negative temperature parameter to balance critic exploitation with behavior cloning.

\section{Discrete Diffusion Skills}
\subsection{Self-Supervised Skill Learning in Discrete Skill Space}

\begin{figure*}[ht]
\vskip 0.2in
\begin{center}
\includegraphics[width=\linewidth]{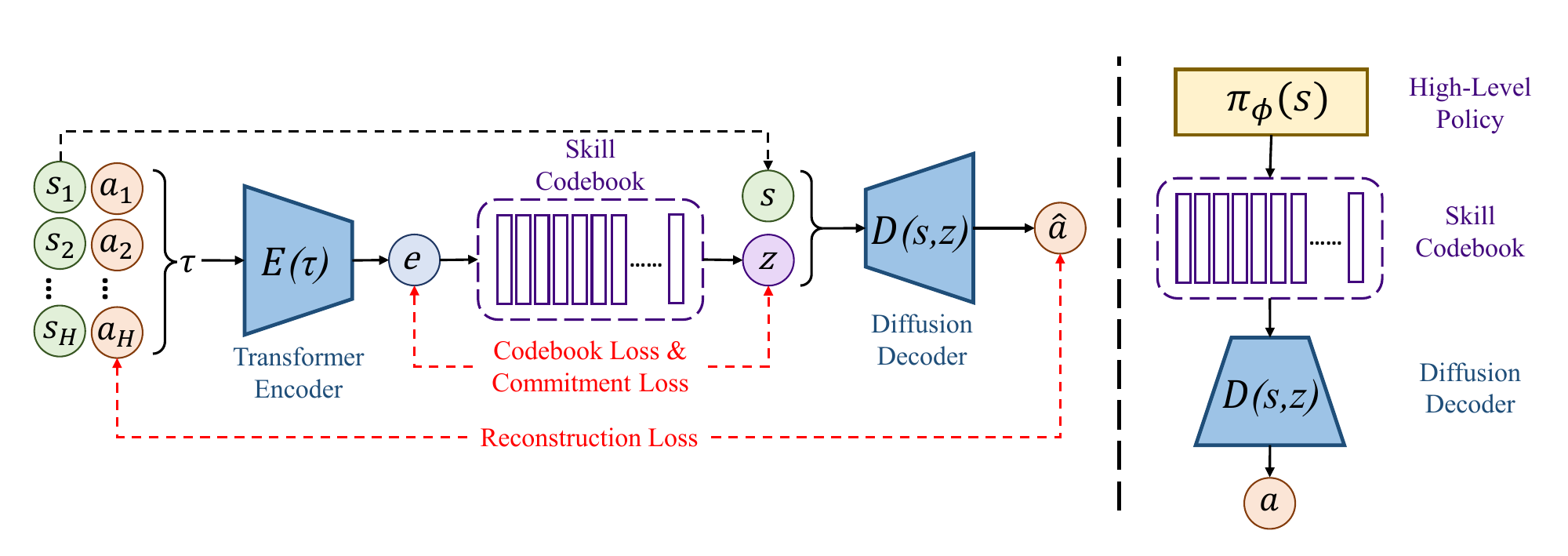}
\vspace{-30pt}
\caption{General framework of Discrete Diffusion Skill. The offline pre-training for discrete skills is illustrated on the left, where the transformer encoder, discrete skill codebook, and diffusion decoder are jointly trained to minimize the codebook loss, the commitment loss, and the reconstruction loss. The online inference is shown on the right. Every $H$ steps, the high-level policy, trained using IQL and AWR, is called to select a skill index. This index is then used to retrieve the corresponding skill vector from the codebook. Subsequently, the action is generated at each time-step using the pre-trained diffusion decoder.}
\label{framework}
\end{center}
\vskip -0.2in
\end{figure*}

To extract meaningful skills from offline datasets in the discrete skill space, we use a skill extraction module based on a VQ-VAE-like structure. This module consists of an encoder that maps state-action sequences to discrete latent skills, and a decoder that reconstructs actions from these discrete skills and corresponding states. The discrete skills are stored in a codebook, denoted as $z_{k}\in R^{D_{z}},$ where $ k\in 1...K$, which is shared between the encoder and the decoder. 

Specifically, the encoder performs a non-linear mapping from a state-action sequence of length $H$, $\tau_{H}=\left \{ s_{0}, a_{0}, s_{1}, a_{1},...,s_{H}, a_{H} \right \}$, to a vector $E(\tau)$. This vector is then quantized by selecting the closest skill vector from the codebook, where each vector $E(\tau)$ is replaced by the content of the nearest skill vector in the codebook:
\begin{align}
Quantize(E(\tau))=z_{k}, \label{eq:quan1}\\
k = \min_{j} ||E(\tau)-z_{j}||. \label{eq:quan2}
\end{align}

The decoder takes the quantized skill vector and the corresponding states to reconstruct the actions via another non-linear function:

\begin{equation}
\hat{a_{t}}=D(z,s_{t}), t\in [1,T]. \label{eq:recon}
\end{equation}

To learn these mappings, the gradient of the reconstruction error is back-propagated through the decoder and the encoder using the straight-through gradient estimator. The skill extraction module includes two additional terms in its objective to align the vector space of the codebook with the encoder's output. The codebook loss, which applies only to the codebook variables, brings the selected skill $z$ closer to the encoder output, $E(\tau)$. The commitment loss, which affects only the encoder weights, encourages the encoder output to remain close to the chosen skill, preventing frequent fluctuations between different code vectors. Thus, the overall objective can be formulated as follows:

\begin{align}
    \mathcal{L}_{skill(\tau_{H})} = &\sum_{H}^{i=1} ||a_{i} - D(z, s_{i})||_{2} 
    + ||\operatorname{sg}[E(\tau_{H})] - z||_{2} \notag \\
    &+ \beta ||\operatorname{sg}[z] - E(\tau_{H})||_{2}, \label{eq:skill_loss}
\end{align}

where $z$ is the quantized skill for the training example $\tau_{H}$, $E$ is the encoder function, and $D$ is the decoder function. The operator $sg$ refers to a stop-gradient operation that blocks gradients from flowing into its argument, and $\beta$ is a hyperparameter that controls the reluctance to change the skill corresponding to the encoder output. 

In practice, we use a transformer encoder to parameterize the encoding process. The offline training datasets are divided into state-action transition sequences of fixed length $H$. The transformer encoder maps each sequence to an embedding vector of dimension $D_{z}$, which is subsequently quantized using equations \ref{eq:quan1} and \ref{eq:quan2}. The codebook is parameterized with standard vector quantization \cite{van2017neural} which has $K$ distinct discrete vector embeddings, each of dimension $D_{z}$ . Our proposed framework demonstrates robustness across different values of $H$, $K$ and $D_{z}$, as confirmed by the ablation study. A diffusion model is selected for the decoder to effectively capture the rich interactions between skill embeddings and states. We employ a VP $\beta$-schedule \cite{song2020score} for the diffusion noise schedule, and a layer-normalized MLP for the noise prediction network, as these components have been empirically shown to be effective in previous work \cite{hansen2023idql} \cite{wang2022diffusion}. Therefore, the loss function \ref{eq:skill_loss} interacts with \ref{eq:diffloss} to define the practical optimization target:
\begin{align}
    \mathcal{L}_{skill(\tau_{H})} = &\mathbb{E}_{t\sim [1,T],x_{0}\sim q(x_{0}),\epsilon \sim \mathcal{N(\mathrm{0},\mathrm{I})}}||\epsilon -\epsilon _{\psi }(x_{t},t )||_{2} \notag \\
    &+ ||\operatorname{sg}[E(\tau_{H})] - z||_{2}
    + \beta ||\operatorname{sg}[z] - E(\tau_{H})||_{2}. \label{eq:skill_diff_loss}
\end{align}
The training process is illustrated in the left part of Figure \ref{framework}.

\subsection{Relabel Offline Datasets with Learned Skills}
After training the transformer encoder, the skill codebook, and the diffusion decoder, the original offline RL dataset is relabeled with the learned skills to form the training dataset $D^{H}$ for the high-level policy. The offline RL dataset is divided into sequences of trajectories of length $H$, $\tau_{H} = \left \{ s_{t}, a_{t}, r_{t}, s^{'}_{t} \right \}_{t\in[1,H]} \sim D$. To transform one of these sequences into a transition step for the high-level policy dataset, the following steps are performed. First, the state at the first time step $s_{1}$ and the next state of the last time step $s^{'}_{H}$ in the snippet are directly used as the state and next state in the high-level policy dataset. Next, a skill vector is inferred based on the state-action pairs in the snippet using the learned encoder. The index of the inferred skill, $k = \min_{j} ||E(\tau)-z_{j}||$, is treated as action in $D^{H}$. Finally, the sum of discounted rewards in the snippet, ${\textstyle \sum_{i=1}^{H}}\gamma^{i-1}r_{i}$, forms the reward of $D^{H}$.

\subsection{IQL for Offline Q-learning on Relabeled Datasets}
To effectively combine pre-trained skills and enable the RL agent to perform complex long-horizon tasks, the relabeled offline RL dataset $D^{H}$ is used to train a high-level policy that predicts the skill to execute based on the state, $\pi_{\phi }(z|s)$. Specifically, the high-level policy first predicts the index of the skill in the codebook, and then the corresponding skill vector is extracted from the codebook.

The Q-function $Q_{\theta}(s,a)$, and the value network $V_{\varphi }(s) $ are trained with equation \ref{eq:iql1} and \ref{eq:iqlloss} respectively. After training the Q-function and value network, the policy is extracted using Advantage Weighted Regression (AWR) \cite{peng2019advantage}. Although the discrete skill space means that implementing the policy is not strictly necessary, we find that the policy extracted via AWR generally improves performance across multiple environments. A possible reason for this is that selecting the action with the highest value is prone to extrapolation error, whereas AWR constrains the policy to focus on data near the distribution, thus improving robustness. 

\subsection{Online Inference with Discrete Skills}
During evaluation, the trained DDS agent selects a skill every $H$ steps using the high-level policy, based on the current state. This matches the horizon used for offline skill training and dataset relabeling. The low-level policy, implemented with a diffusion model, then recursively denoises a random Gaussian prior to generate actions conditioned on the state and the selected skill vector. The number of diffusion steps is set to 5, as this strikes a balance between inference time and sampling quality, as verified in previous works \cite{hansen2023idql}, \cite{wang2022diffusion}. The inference process is illustrated in the right part of Figure \ref{framework}.

\begin{table*}[t]
\caption{Comparison of different offline RL methods using the D4RL benchmarks \cite{fu2020d4rl}. The results for DDS represent the mean and standard deviation of normalized scores over 5 random seeds. DDS demonstrates competitive performance on Locomotion tasks and matches or outperforms baseline methods on long-horizon tasks, including the AntMaze and Kitchen environments.}
\label{offlinerl}
\vskip 0.15in
\begin{center}
\begin{small}
\begin{sc}
\vspace{-20pt}
\resizebox{\textwidth}{!}{
\begin{tabular}{lcccccccccc}
\toprule
Tasks & TD3+BC & IQL & CQL & IDQL & DT & DD & LDCQ & \textbf{DDS(Ours)} \\
\midrule
halfcheetah-m-v2    & 48.3 & 47.4 & 44.0 & \textbf{51.0} & 42.6 & 49.1 & 42.8 & 42.5$\pm$2.6 \\
hopper-m-v2 & 59.3 & 66.3 & 58.5 & 65.4 & 67.6 & 79.3 & 66.2 & \textbf{84.7$\pm$2.1} \\
walker2d-m-v2    & 83.7 & 78.3 & 72.5 & \textbf{82.5} & 74.0 & \textbf{82.5} & 69.4 & 75.1$\pm$3.1 \\
halfcheetah-m-r-v2    & 44.6 & 44.2 & \textbf{47.8} & \textbf{45.9} & 36.6 & 39.3 & 41.8 & 41.6$\pm$1.5 \\
hopper-m-r-v2    & 60.9 & 94.7 & 95.0 & 92.1 & 82.7 & \textbf{100.0} & 66.2 & 84.6$\pm$1.1 \\
walker2d-m-r-v2     & 81.8 & 73.9 & 77.2 & 85.1 & 66.6 & 75.0 & 68.5 & 72.9$\pm$0.4 \\
halfcheetah-m-e-v2  & 90.7 & 86.7 & 91.6 & \textbf{95.9} & 86.8 & 90.6 & 90.2 & \textbf{94.2$\pm$1.6} \\
hopper-m-e-v2  & 98.0 & 91.5 & 105.4 & 108.6 & 107.6 & \textbf{111.8} & \textbf{111.3} & \textbf{109.8$\pm$1.5} \\
walker2d-m-e-v2  & \textbf{110.1} & 109.6 & 108.8 & \textbf{112.7} & 108.1 & 108.8 & 109.3 & 108.3$\pm$0.5 \\
\midrule
locomotion-v2 average & 72.3 & 77.0 & 77.6 & 82.1 & 74.7 & 82.0 & 78.4 & 79.3 \\
\bottomrule
\end{tabular}
}
\resizebox{\textwidth}{!}{
\begin{tabular}{lcccccccccc}
\toprule
Tasks & IQL & CQL & DQL & IDQL & DT & LDCQ & OPAL & \textbf{DDS(Ours)} \\
\midrule
antmaze-u-v2  & 87.5 & 74.0 & \textbf{93.4} & \textbf{94.0} & 59.2 & - & - & \textbf{95.6$\pm$1.2} \\
antmaze-u-d-v2   & 62.2 & 84.0 & 66.2 & 80.2 & 53.0 & - & - & \textbf{94.5$\pm$0.6} \\
antmaze-m-p-v2     & 71.2 & 61.2 & 76.6 & 84.5 & 0.0 & - & - & \textbf{88.5$\pm$0.8} \\
antmaze-m-d-v2     & 70.0 & 53.7 & 78.6 & 84.8 & 0.0 & 75.6 & 81.1 & \textbf{89.1$\pm$0.5} \\
antmaze-l-p-v2      & 39.6 & 15.8 & 46.4 & 63.5 & 0.0 & - & - & \textbf{81.9$\pm$0.8} \\
antmaze-l-d-v2      & 47.5 & 14.9 & 56.6 & 67.9 & 0.0 & 73.6 & 70.3 & \textbf{82.6$\pm$0.6} \\
\midrule
antmaze-v2 average & 63.0 & 50.6 & 69.6 & 79.2 & 18.7 & 74.6 & 75.7 & \textbf{88.7} \\
\bottomrule
\end{tabular}
}
\resizebox{\textwidth}{!}{
\begin{tabular}{lcccccccccc}
\toprule
Tasks & BC & IQL & CQL & DQL & DT & DD & LDCQ & \textbf{DDS(Ours)} \\
\midrule
kitchen-partial-v0   & 38.0 & 46.3 & 50.1 & 60.5 & 42.0 & 57.0 & \textbf{67.8} & \textbf{65.1$\pm$1.5} \\
kitchen-mixed-v0   & 51.5 & 51.0 & 52.4 & 62.6 & 50.7 & \textbf{65.0} & 62.3 & \textbf{67.5$\pm$0.8} \\
\midrule
kitchen-v0 average & 44.8 & 48.7 & 51.3 & 61.6 & 46.4 & 61.0 & 65.0 & \textbf{66.3}\\
\bottomrule
\end{tabular}
}
\end{sc}
\end{small}
\end{center}
\vskip -0.1in
\end{table*}

\section{Experiments}
We conducted a series of quantitative and qualitative experiments with the proposed DDS framework to address the following questions: (1) Is DDS an effective offline RL algorithm?
(2) Do discrete skills provide better interpretability?
(3) Can discrete skills learned from an offline dataset facilitate online learning? (4) How do the skill space parameters, the horizon $H$, and the network architectures impact the performance of DDS?

\begin{figure*}[ht]
\begin{center}
\includegraphics[width=\linewidth]{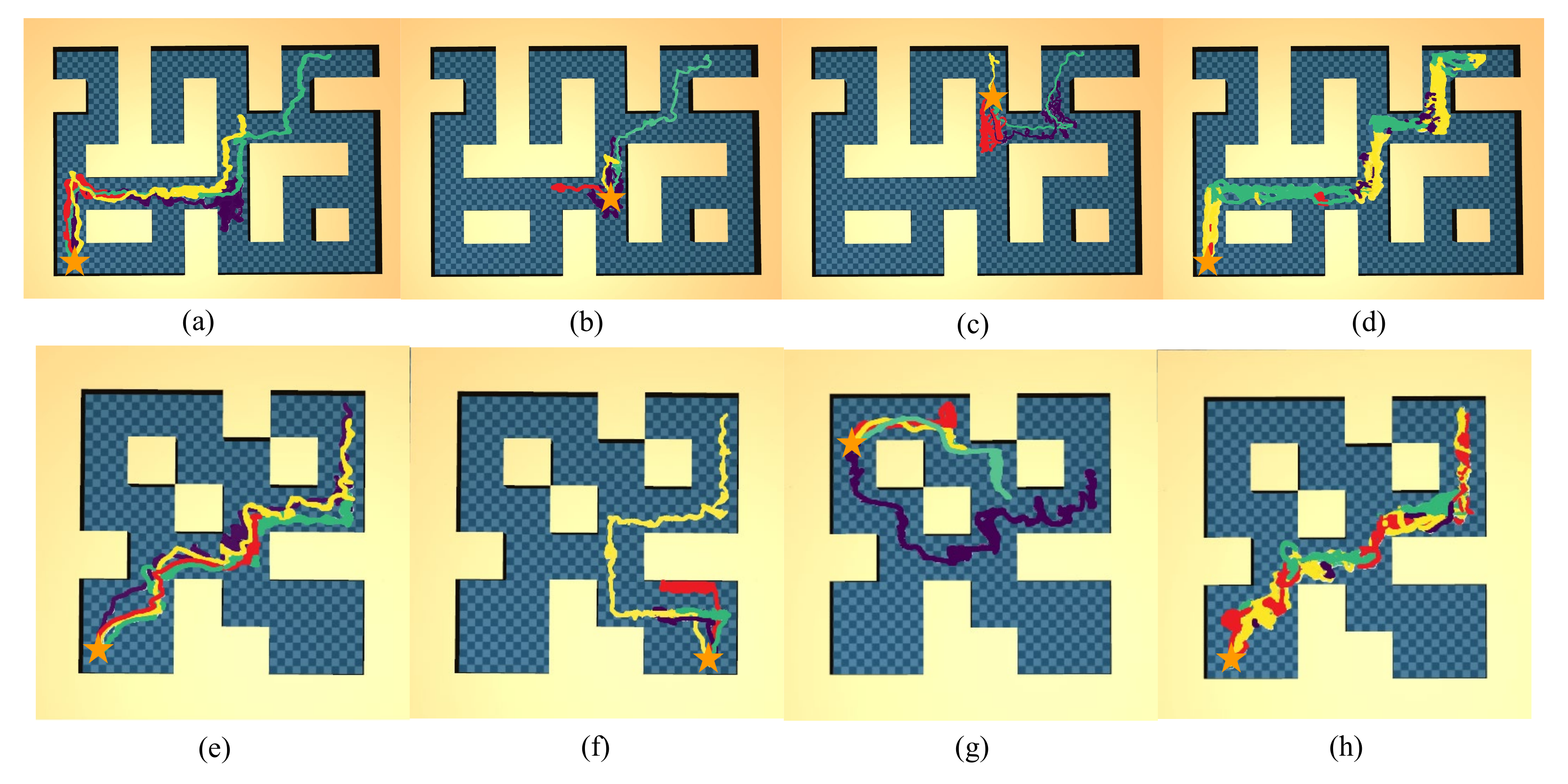}
\vspace{-30pt} 
\caption{Antmaze experiment results: sub-figure (a), (b) and (c) show the trajectories of different skills in Antmaze-Large starting from different positions and sub-figure (d) shows the skill selected in several successful episodes. Sub-figure (e), (f) and (g) show the trajectories of different skills in Antmaze-Medium starting from different positions and sub-figure (h) shows the skill selected in several successful episodes. Different skills are represented with different colors and starting points are denoted with orange stars.}
\label{antmaze}
\end{center}
\vspace{-10pt}
\end{figure*}

\subsection{Offline RL benchmarks}

To demonstrate that our method is a capable offline RL approach, we evaluated it on the widely used D4RL benchmark \cite{fu2020d4rl} and compared it with other offline RL methods.

We consider three distinct task domains from the D4RL benchmark: Gym, AntMaze, and Kitchen. The Gym Locomotion tasks consist of three robot motion control tasks, which are relatively straightforward due to their dense and smooth reward functions. AntMaze presents more challenging tasks with sparse rewards, where the agent must stitch together various suboptimal trajectories to find a path to the goal. The Kitchen environment requires the agent to complete four subtasks to reach a desired state, making long-term value optimization critical. For all environments, we set the number of skills $K=16$, skill dimension $D_{z}=128$ and horizon $H=10$, although other parameter configurations may yield better results for certain environments.

We compare our methods with different classes of baselines. For standard offline RL algorithms, we include the classic BC, TD3+BC \cite{fujimoto2021minimalist}, IQL \cite{kostrikov2021offline}, and CQL \cite{kumar2020conservative}. For diffusion-based offline RL, we consider Diffusion-Q Learning\cite{wang2022diffusion} and IDQL \cite{hansen2023idql}. For sequence modeling approaches, we compare with DT \cite{chen2021decision} and DD \cite{ajay2022conditional}. For skill-based methods, we compare LDCQ \cite{venkatraman2023reasoning} and OPAL \cite{ajay2020opal}. We report the performance of the baseline methods using the best results reported.

The experimental results presented in Table \ref{offlinerl} reveals that our proposed method demonstrates competitive performance on Locomotion tasks. However, its performance degrades when the quality of the data deteriorates. This is likely due to skill pre-training being performed with behavior cloning as the objective, which may result in pre-trained skills containing a significant proportion of suboptimal actions. As the result, this can negatively impact the subsequent training of the high-level policy. Similarly, LDCQ \cite{venkatraman2023reasoning}, another skill-based method, exhibits a similar decrease in performance under degraded data quality, though it is more heavily affected than DDS.

In tasks that require long-term planning, such as the AntMaze and Kitchen environments, our method achieves state-of-the-art performance, particularly excelling in the more challenging AntMaze environment, where it significantly outperforms other baseline methods. This superior performance can be attributed to two key factors: (1) the combination of the trained discrete skill space and the powerful diffusion decoder, which enables precise action reconstruction; and (2) the discretization of actions, which allows the higher-level policy to effectively select skill combinations that maximize the success rate within the discrete skill space.

\begin{table*}[t]
\caption{Success rate of Antmaze Large Diverse task with different skill space settings. DDS demonstrates robust performance across a wide range of parameter configurations. Results are presented as the mean $\pm$ standard deviation of normalized scores over 5 random seeds.}
\vspace{-10pt}
\label{ablation}
\vskip 0.15in
\begin{center}
\begin{small}
\begin{sc}
\begin{tabular}{cccccc}
\toprule
\multirow{2}{*}{\textbf{Skill Dimension}} & \multicolumn{4}{c}{\textbf{Number of Skills}} \\
\cmidrule(lr){2-5}
 & \textbf{4} & \textbf{8} & \textbf{16} & \textbf{32} \\
\midrule
\textbf{32}  & 85.2 $\pm$ 0.2 & 85.1 $\pm$ 1.5 & 62.8 $\pm$ 0.9 & 0.0 \\
\textbf{64}  & 79.3 $\pm$ 0.5 & 88.9 $\pm$ 0.3 & 68.3 $\pm$ 0.4 & 0.0 \\
\textbf{128} & 76.3 $\pm$ 0.7 & 77.8 $\pm$ 0.3 & 82.6 $\pm$ 0.6 & 78.2 $\pm$ 0.2 \\
\textbf{256} & 81.5 $\pm$ 0.3 & 85.2 $\pm$ 0.2 & 81.6 $\pm$ 1.1 & 76.8 $\pm$ 0.7 \\
\bottomrule
\end{tabular}
\end{sc}
\end{small}
\end{center}
\vskip -0.1in
\end{table*}

\subsection{Interpretability of Discrete Skills} 
We present results from the Hopper, AntMaze-Large, and Antmaze-Medium environments \cite{fu2020d4rl} to demonstrate that pre-trained discrete skill embeddings exhibit distinct and interpretable meanings. Using D4RL \cite{fu2020d4rl} offline datasets, we train discrete skills for all three environments. In this section, we set the number of skills $K=4$, the dimension of the skill vector $D_{z}=64$, and the horizon $H=10$.  

\vspace{-10pt} 
\begin{figure}[H]
\begin{center}
\includegraphics[width=\columnwidth]{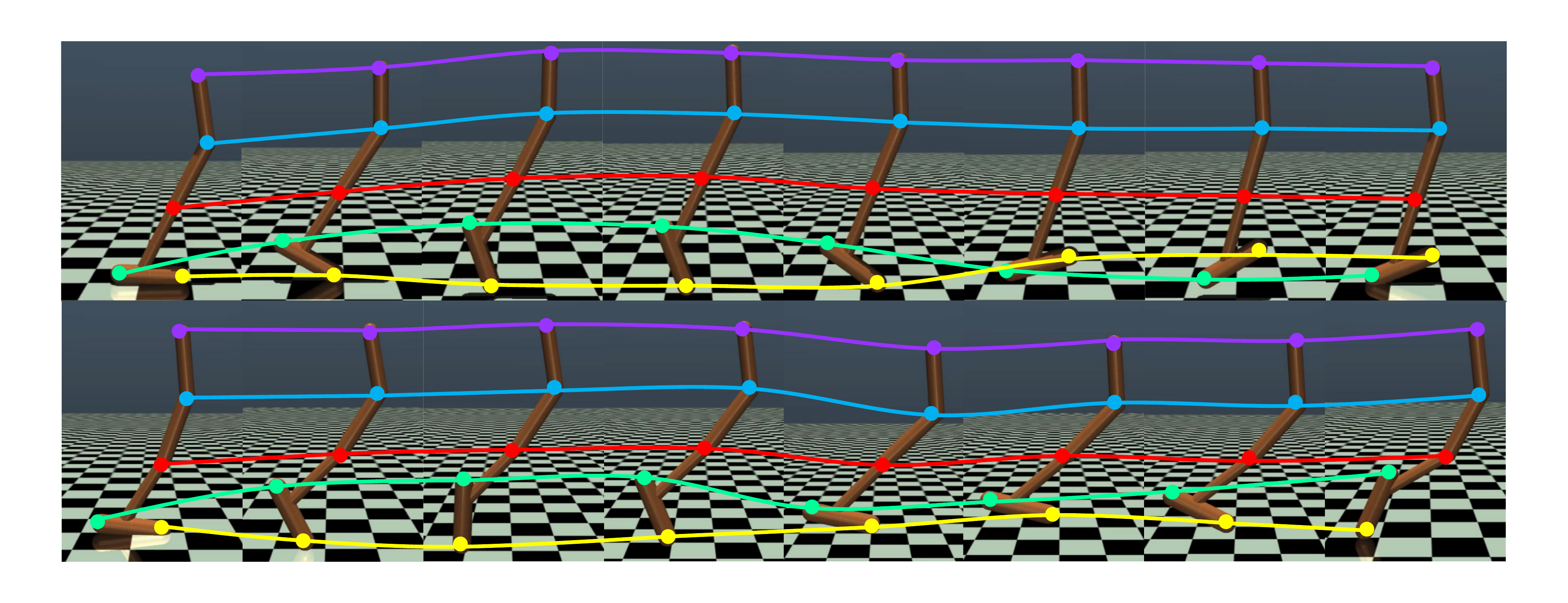}
\vspace{-30pt} 
\caption{The the movement of the hopper using two different skills: the upper graph shows smoother motion, while the lower one exhibits more aggressive movement. Joints are connected between frames to illustrate moving pattern.}
\label{hopper}
\end{center}
\end{figure}
\vspace{-20pt}

We replay various skills learned from the Hopper-Medium-Expert-v2 dataset to illustrate that different skills represents distinct behaviors. As shown in Figure \ref{hopper}, movement patterns of joints differ significantly across skills. This is further validated in the AntMaze-Large and Antmaze-Medium environment, as demonstrated in Figure \ref{antmaze}. We replay different skills from different starting locations and show the trajectories generated by different skills with different colors. It is obvious that the agent’s behavior varies with different skills. Additionally, it is important to note that the discrete skills interact richly with the state, as the same skill can exhibit completely different behaviors depending on the agent's position. 

To demonstrate the skill stitching achieved by the high-level policy, we record the trajectory and selected skill for several successful episodes in the AntMaze environments. As shown in sub-figure (d) and (h) of Figure \ref{antmaze}, different skills are selected at different stages of an episode to maximize the success rate.

It is worth noting that some skills may have the capability to complete the entire task. For example, the skill represented by the green lines in sub-figure (a) of Figure \ref{antmaze}, which could accomplish the task on its own. However, the acquisition of such a skill is not guaranteed and the success rate of replaying a single skill is relative low. Therefore, it is evident that the trained high-level policy learns to select different skills based on the circumstances to maximize the probability of task completion, rather than collapsing to a single skill.

Through visual analysis of the discrete skill space, we observe that the discrete skill space offers better interpretability compared to the continuous skill space. This enhanced interpretability makes the discrete skill space particularly suitable for applications requiring high levels of transparency and safety, such as robotic control and autonomous driving.

\subsection{Online RL with Pre-Trained Discrete Skills}
\begin{figure}[H]
\begin{center}
\includegraphics[width=\columnwidth]{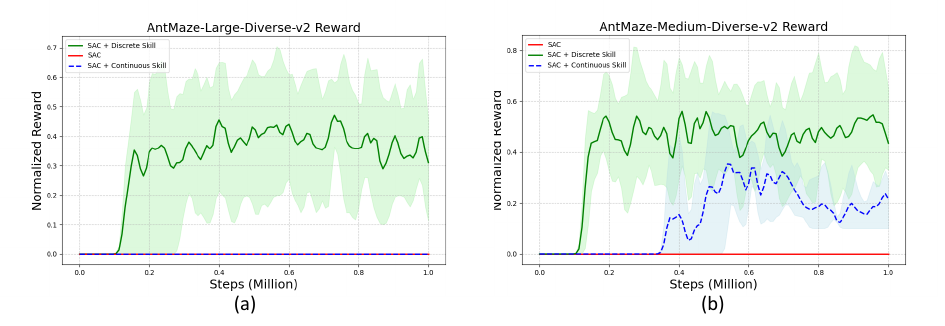}
\vspace{-30pt} 
\caption{Online training results with and without pre-trained discret and continuous skills in AntMaze-Large-Diverse-v2 and AntMaze-Medium-Diverse-v2}
\label{online}
\end{center}
\end{figure}

In this section, our experiments aim to illustrate how offline pre-trained discrete skills can enhance exploration, thereby facilitating online RL. We conducted experiments in the challenging AntMaze-Large-Diverse and AntMaze-Medium-Diverse environments. We first train skills in both discrete and continuous spaces using identical offline datasets. These trained skills effectively redefine the action space available to the high-level policy, thereby simplifying the online training process with the SAC algorithm \cite{haarnoja2018soft}. As a baseline comparison, we also trained an agent directly in the original action space of the environments using the SAC algorithm with identical configurations. Specific parameters could be found in Appendix \ref{online_param}, the training performance of these agents is illustrated in Figure \ref{online}.

The results demonstrate that the discrete skill space significantly enhances exploration compared to the original action space and the continuous skill space. Agents utilizing discrete skills achieve competitive performance quickly in both environments, navigating the environments more effectively. Conversely, other agents struggle with effective exploration and encounter difficulties in complex, sparse reward tasks.

\begin{table*}[ht] 
\caption{Success rate of Antmaze Large Diverse task with different network architecture combinations. Our transformer encoder and diffusion decoder lead other combinations by a large margin. Results are presented as the mean $\pm$ standard deviation of normalized scores over 5 random seeds.}
\label{table:network}
\begin{center}
\begin{small}
\vspace{-10pt}
\resizebox{\textwidth}{!}{
\begin{tabular}{lcccc}
\toprule
Parameter & GRU + MLP & Transformer + MLP & GRU + Diffusion &  Transformer + Diffusion \\
\midrule
AntMaze-Large-Diverse & 43.4$\pm$2.4 & 63.0$\pm$0.8 & 32.2$\pm$1.3 & 82.6$\pm$0.6 \\
Kitchen-Mixed & 50.3$\pm$2.7 & 51.4$\pm$1.8 & 31.3$\pm$3.2 & 67.5$\pm$0.8 \\
\bottomrule
\end{tabular}
}
\end{small}
\end{center}
\vskip -0.1in
\end{table*}

The results of this experiment demonstrate that the discrete skill space can assist online RL agents with exploration, improving their performance in long-horizon tasks with sparse rewards, as theoretically proved by \cite{li2024skills}. Furthermore, the discrete skill space facilitates the discretization of continuous action spaces. Previous research \cite{luo2023action} has shown that discrete action spaces are more conducive to the training of reinforcement learning agents.

\vspace{-20pt} 

\subsection{Ablation Studies}

\textbf{Skill Space Parameters} We begin by conducting ablation studies on the number of skills and the dimensionality of the skill vectors. These experiments are performed in the AntMaze-Large-Diverse environment to assess how variations in skill numbers and dimensions impact the agent's performance. Our results show that DDS exhibits strong robustness to the dimensionality of the discrete skill space, achieving good performance across a wide range of skill space configurations.

Analyzing the experimental data in Table 2, we observe that when the skill dimensionality is large, the agent's performance is relatively insensitive to the number of skills. However, when the skill dimensionality is small, the number of skills significantly affects the agent's performance. Through qualitative analysis, we find that with a larger skill dimensionality, each skill captures more information, allowing even a small number of skills to represent rich trajectory details. In this case, increasing the number of skills can lead to some skills becoming similar and exhibiting similar behaviors. This essentially results in an underutilized skill space but it does not negatively affect the agent's performance.

In contrast, when the skill dimensionality is small, each skill contains less information. As the number of skills increases, the behaviors represented by each skill become more distinct. With many diverse skills, training the high-level policy using IQL becomes more challenging, leading to local optima and negatively affecting the agent’s performance. Additionally, we find that aligning the number of skills with the dimensionality of the skill vectors, i.e., using parameter configurations near the diagonal in Table \ref{ablation} benefits overall performance.

\textbf{Network Architectures} In addition to conducting ablation studies on the configuration of the skill space, we also performed ablation experiments on the encoder and decoder network architectures. We compare GRU-based encoder and MLP-based decoder against our proposed transformer encoder and diffusion decoder in Kitchen-Mixed-v0 and AntMaze-Large-Diverse-v2 environment. The GRU-based encoder has a structure similar to that in \cite{venkatraman2023reasoning} and a similar amount of parameters to our transformer encoder. The MLP-based decoder uses the same backbone network as our diffusion decoder. 

As the results in Table \ref{table:network} suggest, our proposed network structures lead previous networks by a large margin thanks to their excellent ability to model complex distributions. 

\vspace{-30pt} 
\begin{figure}[H]
\vskip 0.2in
\begin{center}
\includegraphics[width=\columnwidth]{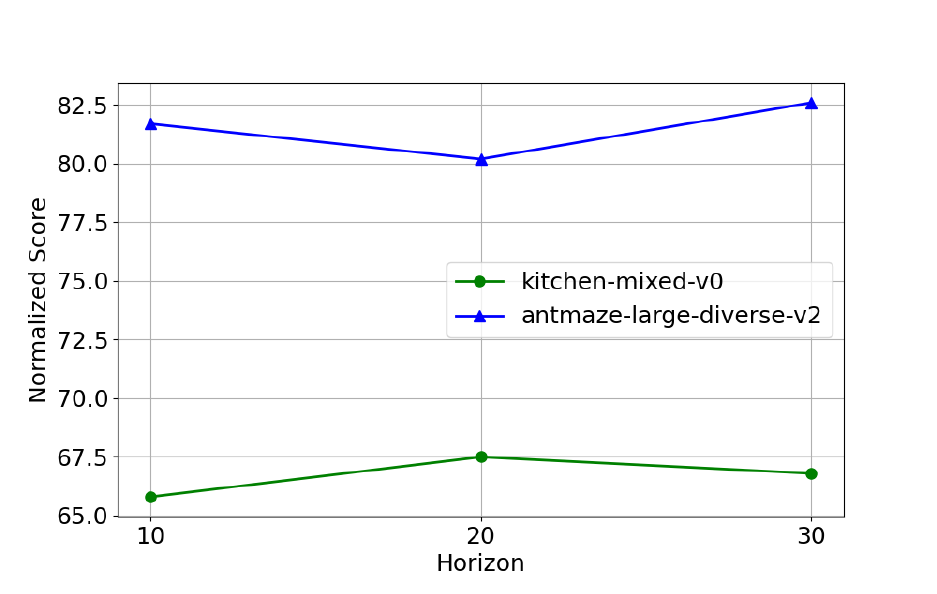}
\vspace{-30pt} 
\caption{Ablation study on the skill horizon $H$.}
\label{horizon}
\end{center}
\vskip -0.2in
\end{figure}

\textbf{Horizon $H$} We performed ablation experiments on the horizon $H$ during skill learning and online inference. These experiments were carried out on the Kitchen-Mixed-v0 and AntMaze-Large-Diverse-v2 datasets. For this section, we fixed the number of skills $K=16$ and the dimension of the skill vector $D_{z}=128$. The results demonstrate that our proposed method is not sensitive to the horizon $H$.

As shown in the results in Figure \ref{horizon}, the horizon $H$ did not have a significant impact on agent performance in either of the test environments. Several factors may explain this phenomenon: first, regardless of how frequently skills are chosen during online inference, the high-level policy trained with the IQL algorithm can effectively stitch together different skills to complete the task; second, increasing the length of the data segment does not lead to an increase in reconstruction error during the skill training process.

\section{Conclusion}
In this work, we apply discrete skill spaces to address long-horizon and sparse reward challenges in offline RL. Through extensive experiments, we demonstrated the effectiveness, interpretability, and robustness of the proposed method across various parameter settings, as well as its advantages for exploration and action space discretization. Given the benefits of discrete action spaces observed in offline RL tasks, we believe this approach can be extended to more complex domains, such as robotic control and autonomous driving.

\section*{Limitations}
Due to the behavior cloning objective in skill extraction, the final performance of the RL agent is correlated with the quality of the offline dataset. And when the number of skills is large, the high-level policy may get stuck in local optima and lead to suboptimal performance of the agent; however, this could be alleviated by carefully selecting the skill space parameters as shown in the ablation study.

\section*{Impact Statement}
Our work enhances the performance and interpretability of skill-based offline RL methods by constructing a discrete skill space. It also improves the stability of high-level policy training in hierarchical RL frameworks and aids agents in better exploration.


\nocite{langley00}

\bibliography{example_paper}
\bibliographystyle{icml2025}

\newpage
\appendix
\section{Experiment Details}
\subsection{Network Architecture}
\subsubsection{Transformer Encoder}
The transformer encoder module includes:
\begin{itemize}
    \item Embedding Layer: A linear layer that projects concatenated state-action inputs into the hidden dimension.
    \item Transformer layers: Transformer layers to model temporal dependencies in the sequence.
    \item Pooling Layer: An adaptive average pooling layer to aggregate the sequence into a single representation.
    \item Projection Layer: A two-layer MLP (with ReLU activation) that maps the pooled features into the skill dimension.
\end{itemize}

The state and action sequence are concatenated along the last dimension to form a single tensor. The tensor is fed through the embedding layer to map the concatenated sequence into a tensor in hidden dimension. Learnable positional encoding is added to the embedded sequence to inject temporal information. Passes the sequence through the Transformer encoder to capture temporal dependencies and interactions. Applies adaptive average pooling across the sequence dimension, reducing the time dimension. Maps the pooled features to the skill dimension using the projection layer.

\begin{table}[H]
\caption{Hyperparameter values for Transformer Encoder.}
\label{table:parameter_tr_en}
\vskip 0.15in
\begin{center}
\begin{small}
\begin{tabular}{lc}
\toprule
Parameter & Value \\
\midrule
Transformer Layers & 4 \\
Head Number & 8 \\
Dropout & 0.1 \\
Hidden Dimension & 256 \\
\bottomrule
\end{tabular}
\end{small}
\end{center}
\vskip -0.1in
\end{table}

\subsubsection{Diffusion Decoder}
The diffusion decoder module uses VP-$\beta$ schedule \cite{song2020score} and the noise prediction network is parameterized with a layer-norm MLP which has:
\begin{itemize}
    \item Sinusoidal Time Embedding: Transformed time input into a high-dimensional representation.
    \item Input Projection layer: A fully connected layer that project the concatenated input into the hidden dimension.
    \item A Stack of Dense Fully Connected Layer Blocks: Each blocks contains 2 linear layers which first expand the input dimensionality by 4 times and then compresses back to the input dimensionality, blocks also have residual connection and layer normalization. 
    \item Output Projection layer: A fully connected layer that maps the pooled features into the skill dimension.
\end{itemize}

\begin{table}[H]
\caption{Hyperparameter values for Diffusion Decoder.}
\label{table:parameter_diff_de}
\vskip 0.15in
\begin{center}
\begin{small}
\begin{tabular}{lc}
\toprule
Parameter & Value \\
\midrule
Diffusion Steps & 5 \\
$\beta_{min}$ & 0.1 \\
$\beta_{max}$ & 10 \\
Dropout & 0.1 \\
Hidden Dimension & 256 \\
Time Embedding Dimension & 16 \\
Number of Dense Fully Connected Layer Blocks  & 4 \\
\bottomrule
\end{tabular}
\end{small}
\end{center}
\vskip -0.1in
\end{table}

\subsubsection{Value Network}
Value networks is modeled with a MLP, which has 2 hidden layers with ReLU activations and 256 hidden units for all hidden layers.

\subsubsection{Q-Function Network}
We build two Q networks with the same MLP, which has 2 hidden layers with ReLU activations and 256 hidden units for all hidden layers.

\subsubsection{Policy Network}
Policy networks is modeled with a MLP, which has 2 hidden layers with ReLU activations and 256 hidden units for all hidden layers.

\subsection{Hyperparameters}

\subsubsection{Skill Training Parameters}
\begin{table}[H]
\caption{Hyperparameter values for skill extraction.}
\label{table:parameter_skill}
\vskip 0.15in
\begin{center}
\begin{small}
\begin{tabular}{lc}
\toprule
Parameter & Value \\
\midrule
Learning Rate & 5e-5 \\
Batch Size & 128 \\
Gradient Steps & 500000 \\
Skill Dimension & 128 \\
Skill Number & 16 \\
$\beta$ & 0.25 \\
Horizon $H$ & 10 \\
Optimizer & Adam \\
\bottomrule
\end{tabular}
\end{small}
\end{center}
\vskip -0.1in
\end{table}

\subsubsection{IQL Training Parameters}
\begin{table}[H]
\caption{Hyperparameter values for IQL.}
\label{table:parameter_iql}
\vskip 0.15in
\begin{center}
\begin{small}
\begin{tabular}{lc}
\toprule
Parameter & Value \\
\midrule
Learning Rate & 1e-4 \\
Discount & 0.99 \\
Batch Size & 256 \\
Gradient Steps (Q-Learning) & 1000000 \\
Gradient Steps (AWR) & 500000 \\
$\tau$ & 0.7 \\
Optimizer & Adam \\
EMA $\alpha$ & 0.005 \\
\bottomrule
\end{tabular}
\end{small}
\end{center}
\vskip -0.1in
\end{table}

\subsubsection{Online Training Parameters} \label{online_param}
\begin{table}[H]  
\caption{Hyperparameter values for SAC and offline pre-training.}
\label{table:parameter_sac}
\vskip 0.15in
\begin{center}
\begin{small}
\begin{tabular}{lc}
\toprule
Parameter & Value \\
\midrule
Learning Rate & 3e-4 \\
Discount & 0.99 \\
Batch Size & 256 \\
Initial Temperature & 0.1 \\
Optimizer & Adam \\
EMA $\alpha$ & 0.005 \\
Continuous Skill Space Dimension & 8 \\
Discrete Skill Dimension & 128 \\
Discrete Skill Number & 32 \\
Horizon $H$ & 10 \\
\bottomrule
\end{tabular}
\end{small}
\end{center}
\vskip -0.1in
\end{table}

\subsection{Hardware}
Most of the experiments were conducted with a single NVIDIA RTX4090. Skill pre-training took approximately 2.5 hours, and high-level policy training took about 3 hours.


\end{document}